\documentclass[review]{elsarticle}

\usepackage{lineno,hyperref}

\usepackage{algorithm}
\usepackage{algorithmic}
\usepackage{amsfonts}
\usepackage{multirow}
\usepackage{subfig}
\usepackage{amssymb,amsmath}
\usepackage{booktabs}
\usepackage{setspace}
\usepackage{dsfont}

\usepackage{caption} % for color of caption
\usepackage{xcolor} % for color of caption
\usepackage{booktabs} % for color of table
\usepackage{colortbl} % for color of table

\journal{Journal of \LaTeX\ Templates}

\begin{document}

\begin{frontmatter}

\title{Deep Momentum Uncertainty Hashing}

% include affiliations in footnotes:
\author[address1,address3,address4]{Chaoyou Fu\corref{authorinformation}}
\cortext[authorinformation]{This work was done during an internship at Horizon Robotics.}
\ead{chaoyou.fu@nlpr.ia.ac.cn}

\author[address5]{Guoli Wang}
\ead{wangguoli1990@mail.tsinghua.edu.cn}

\author[address3]{Xiang Wu}
\ead{alfredxiangwu@gmail.com}

\author[address6]{Qian Zhang}
\ead{qian01.zhang@horizon.ai}

\author[address1,address2,address3,address4]{Ran He\corref{correspondingauthor}}
\cortext[correspondingauthor]{Corresponding author.}
\ead{rhe@nlpr.ia.ac.cn}

\address[address1]{National Laboratory of Pattern Recognition, CASIA}
\address[address2]{Center for Excellence in Brain Science and Intelligence Technology, CAS}
\address[address3]{Center for Research on Intelligent Perception and Computing, CASIA}
\address[address4]{School of Artificial Intelligence, University of Chinese Academy of Sciences}
\address[address5]{Department of Automation, Tsinghua University}
\address[address6]{Horizon Robotics}

\begin{abstract}
Combinatorial optimization (CO) has been a hot research topic because of its theoretic and practical importance.
As a classic CO problem, deep hashing aims to find an optimal code for each data from finite discrete possibilities, while the discrete nature brings a big challenge to the optimization process.
Previous methods usually mitigate this challenge by binary approximation, substituting binary codes for real-values via activation functions or regularizations. However, such approximation leads to uncertainty between real-values and binary ones, degrading retrieval performance. In this paper, we propose a novel Deep Momentum Uncertainty Hashing (DMUH). It explicitly estimates the uncertainty during training and leverages the uncertainty information to guide the approximation process. Specifically, we model \emph{bit-level uncertainty} via measuring the discrepancy between the output of a hashing network and that of a momentum-updated network. The discrepancy of each bit indicates the uncertainty of the hashing network to the approximate output of that bit. Meanwhile, the mean discrepancy of all bits in a hashing code can be regarded as \emph{image-level uncertainty}. It embodies the uncertainty of the hashing network to the corresponding input image. The hashing bit and image with higher uncertainty are paid more attention during optimization. To the best of our knowledge, this is the first work to study the uncertainty in hashing bits. Extensive experiments are conducted on four datasets to verify the superiority of our method, including CIFAR-10, NUS-WIDE, MS-COCO, and a million-scale dataset Clothing1M. Our method achieves the best performance on all of the datasets and surpasses existing state-of-the-art methods by a large margin.
\end{abstract}


\end{frontmatter}

%\linenumbers

\section{Introduction}\label{Introduction}
Combinatorial optimization (CO) has a great impact on business and society, ranging from locomotive dispatching to aerospace industry \cite{bengio2020machine,yan2015multi}. 
Due to the limitation in tractability and scalability of traditional solvers, many researchers recently turn their attention to machine learning (ML) for better solutions \cite{vinyals2015pointer,wang2020combinatorial}. 
Deep hashing is a typical task that combines CO and ML, aiming to find an optimal code for each data from finite discrete possibilities via deep neural networks, so that similar data have shorter Hamming distance and dissimilar data have longer Hamming distance.

With the explosive growth of data in practical applications, hashing has received sustained attention due to its advantages in low storage cost and fast computation speed \cite{wang2018survey}.
Traditional hashing methods are based on the elaborately designed hand-crafted features \cite{lin2014fast,he2014robust,he2009robust}. 
The binary codes are learned from data distributions \cite{gong2011iterative} or obtained by random projection \cite{gionis1999similarity}.
In recent years, as the thriving of deep neural networks, deep hashing that combines hashing with deep neural networks further improves retrieval performance \cite{li2015feature}.
Generally, the last layer of a neural network is leveraged to output binary hashing codes \cite{fu2018neurons}.
Early works, such as convolutional neural network hashing (CNNH) \cite{xia2014supervised}, adopt a two-stage framework, where the feature learning of the neural network and the hashing coding are separate.
Subsequent works, \textit{e.g.} deep pairwise-supervised hashing (DPSH) \cite{li2015feature}, perform feature learning and hashing coding in an end-to-end framework, which has shown better performance than the two-stage one.
For all of the deep hashing methods, an intractable problem is that the binary hashing codes are discrete, which impedes the back-propagation of gradient in the neural network \cite{jiang2018deep}.
The discrete optimization of the binary hashing codes remains a great challenge.

\begin{figure}[t]
\centering
\includegraphics[width=0.5\textwidth]{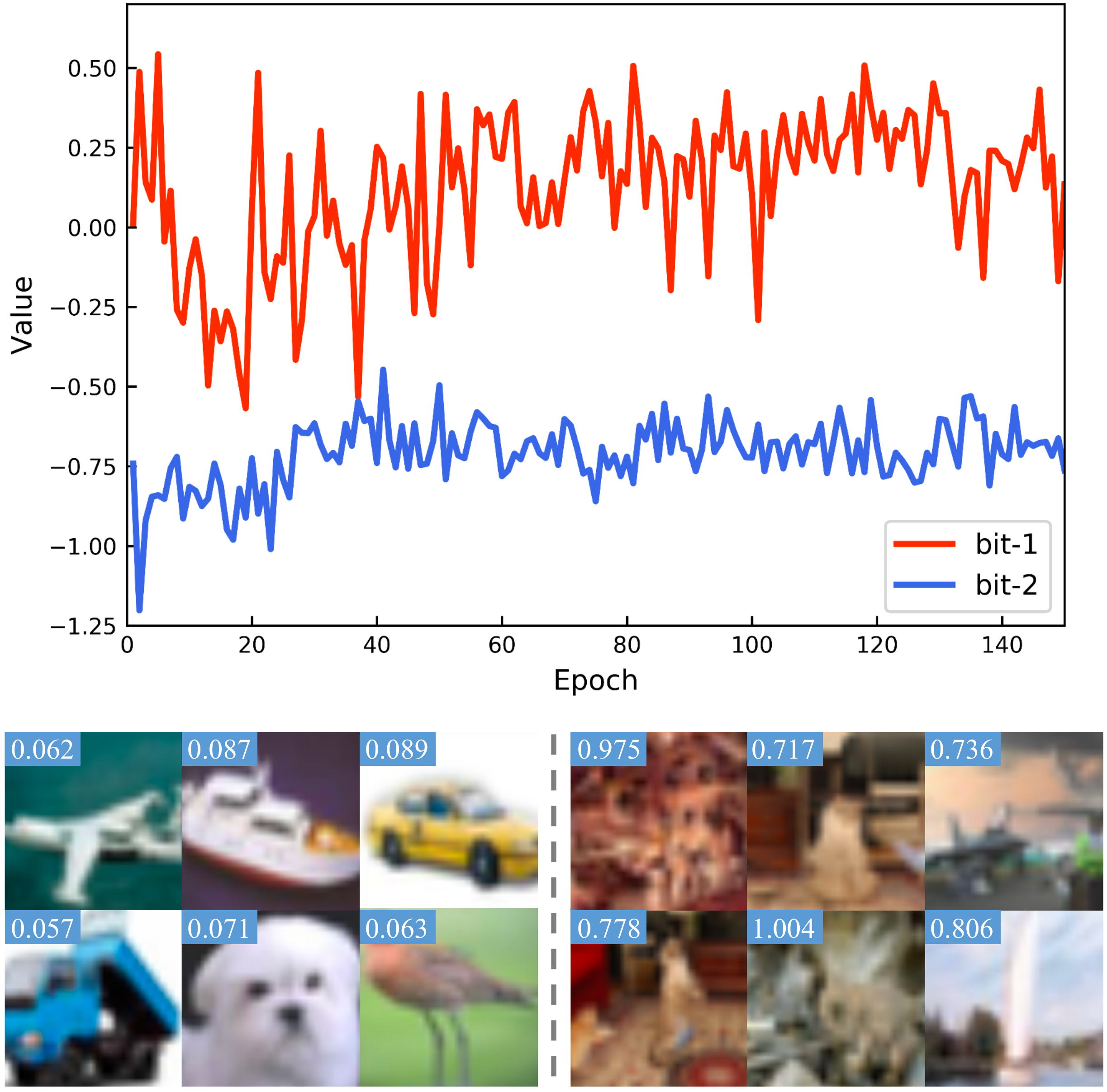}
\caption{\textbf{Top}: the approximate real-values of two bits at different training epochs. 
It is obvious that bit-1 changes more sharply than bit-2, which denotes that the hashing network has higher \emph{bit-level uncertainty} to the former.
\textbf{Bottom}: the mean bit-level uncertainty of all bits in a hashing code is regarded as \textit{image-level uncertainty}, whose value is shown in the upper left corner of each image.
The images with low uncertainty ($<$ 0.1) usually have clear objects and simple backgrounds (left), while the images with high uncertainty ($>$ 0.7) contain nebulous objects and complex scenarios (right).
}
\label{fig-uncertainty}
\end{figure}

Previous methods usually adopt binary approximation to tackle the above challenge.
That is, the binary codes are replaced by continuous real-values, which are enforced to be binary via non-linear activation functions \cite{jiang2018deep}.
Nevertheless, the output of the activation function, such as \emph{Sigmoid} or \emph{Tanh}, is easy to be saturated.
This inevitably slows down or even limits the training process \cite{liu2016deep}.
In order to avoid the saturating problem, some recent methods desert the non-linear activation function and impose a regularization on the output to enforce the real-value of each bit to be close to a binary one (+1 or -1) \cite{li2015feature}.
However, these methods equally approximate all bits, while ignore their differences.
As shown in Fig.~\ref{fig-uncertainty} (Top), we discover that the approximate output of each bit has a unique change trend. 
It is observed that the output of bit-1 has more drastic changes than the output of bit-2 during training. 
That is to say, the hashing network has higher uncertainty to the approximate output of bit-1 than that of bit-2.
We call such uncertainty for each bit as \emph{bit-level uncertainty}.
Furthermore, if all bits of a hashing code generally have high uncertainty, it indicates that the hashing network has high uncertainty to the corresponding input image. 
We define the mean bit-level uncertainty of all bits in a hashing code as the \emph{image-level uncertainty}.
As can be seen from Fig.~\ref{fig-uncertainty} (Bottom), the images with high image-level uncertainty usually contain more complex scenarios, belonging to hard examples \cite{lin2017focal,wu2017sampling}.

In order to explicitly estimate the bit-level uncertainty, \textit{i.e.} the change trends of the hashing bits, it is required to compare current output values with previous ones.
A straightforward strategy is to store the outputs of all training images in each optimization step and then compare the current outputs with them. 
Unfortunately, this strategy is unfeasible because of the requirement of huge storage memory when training on large-scale datasets.
Recently, in order to tackle the memory problem in unsupervised and semi-supervised learning, some works \cite{he2019momentum,tarvainen2017mean,french2017self} develop an extra momentum-updated network that averages model weights during training.
The momentum-updated network is an ensemble of previous networks in different optimization steps, outputting ensemble results \cite{french2017self}. 
Inspired by this, in our method, a momentum-updated network is introduced to obtain previous outputs approximately.
We further compare the outputs between the hashing network and the momentum-updated network, and regard the discrepancy as the bit-level uncertainty.
According to the magnitude of the uncertainty, we set different regularization weights for different hashing bits.
Besides, by averaging the uncertainty of all bits in a hashing code, we obtain the image-level uncertainty of the corresponding input image.
The image with higher uncertainty is paid more attention during the optimization of Hamming distance.
The effectiveness of our method is demonstrated on four challenging datasets, including CIFAR-10 \cite{krizhevsky2009learning}, NUS-WIDE \cite{chua2009nus}, MS-COCO \cite{lin2014microsoft}, and a million-scale dataset Clothing1M \cite{xiao2015learning}.
In summary, the main contributions of this work are as follows:

\begin{itemize}
  \item We are the first to explore the uncertainty of hashing bits during approximate optimization. Depending on the magnitude of uncertainty, the corresponding hashing bits and input images receive different attention.

  \item We propose to explicitly model bit-level and image-level uncertainty, resorting to the output discrepancy between the hashing network and the momentum-updated network. 

  \item Extensive experiments on the CIFAR-10, the NUS-WIDE, the MS-COCO, and the large-scale Clothing1M datasets demonstrate that our method significantly improves retrieval performance when compared with state-of-the-art methods.
\end{itemize}

\section{Related Work}
\subsection{Learning of Combinatorial Optimization}
A growing body of research is dedicated to integrating CO and ML, since the latter can make effective decisions for the former with low computational costs \cite{wang2020combinatorial,bengio2020machine}.
\cite{vinyals2015pointer} develops a Pointer Net that is equipped with neural attention, which has the ability to find an approximate solution for the Travelling Salesman Problem (TSP).
\cite{huang2019coloring} proposes a novel FastColorNet that adopts deep reinforcement learning to color large graphs. 
\cite{wang2019learning,wang2020combinatorial,wang2021neural} focus on leveraging graph embedding networks for robust graph matching \cite{yan2020learning,ma2021image}.
\cite{zaheer2017deep} develops a DeepSets architecture for node sets and also provides corresponding permutation invariant functions.
With respect to resource management in wireless networks, \cite{shen2019lorm} presents a LORM framework that uses imitation learning to find the best pruning policy.
In contrast to the foregoing methods, this paper studies deep hashing that aims to learn the optimal binary code for each data via deep neural networks.
Meanwhile, this paper also explores the uncertainty in the optimization process, which is expected to provide new insights for other combinatorial problems.

\subsection{Hashing Retrieval}
Hashing aims to project data from high-dimensional pixel space into the low-dimensional binary Hamming space \cite{jiang2017asymmetric}.
It has drawn substantial attention of researchers due to the low time and space complexity.
Current hashing methods can be grouped into two categories, including data-independent hashing methods and data-dependent hashing methods.
For the data-independent hashing methods, the binary hashing codes are generated by random projection or manually constructed without using any training data.
Locality sensitive hashing (LSH) \cite{gionis1999similarity} is a representative method.
Since the data-independent hashing methods usually require long code length to guarantee retrieval performance, more efficient data-dependent hashing methods that learn hashing codes from training data have gained more attention in recent years \cite{jiang2017asymmetric}.

The data-dependent hashing methods can be further divided into two types, \textit{i.e.} unsupervised methods and supervised methods, according to whether using the similarity labels.
Iterative quantization hashing (ITQ) \cite{gong2011iterative} and ordinal embedding hashing (OEH) \cite{liu2016towards} are representative unsupervised hashing methods.
Both of them retrieval the neighbors by exploring the metric structure in the data.
Other unsupervised hashing methods include discrete graph hashing (DGH) \cite{liu2014discrete}, inductive manifold hashing (IMH) \cite{shen2013inductive}, and global hashing system (GHS) \cite{tian2016global}. 
Although unsupervised learning avoids the annotation demand of the training data, exploiting available supervisory information usually implies better performance.
Representative supervised hashing methods based on hand-crafted features include supervised hashing with kernels (KSH) \cite{liu2012supervised} and column-sampling based discrete supervised hashing (COSDISH) \cite{kang2016column}, both of which achieve impressive results.

Benefiting from the powerful representation ability of deep neural networks \cite{zhang2018deep}, hashing has made further progress in the last few years \cite{xu2019graph,yang2019distillhash,lu2020adversarial}.
\cite{sun2019supervised} is the first one to introduce cross-modal hashing with hierarchical labels to settle real-world problems and also contributes a large-scale dataset.
\cite{li2019supervised,yan2016supervised} propose to leverage full label information to assist in the learning of multimodal hashing, and present a discrete optimization algorithm to learn binary codes.
\cite{deng2019unsupervised} develops a generative adversarial framework for unsupervised hashing and introduces a semantic similarity matrix to guide hashing coding.
\cite{deng2018triplet} digs similarity correlations between cross-modal data via a triplet sampling strategy, and elaborately designs an objective function to learn discriminative hashing codes.
\cite{li2018deep} builds two subnetworks to learn potential semantic correlations in cross-modal data and hashing codes, respectively.
\cite{li2020vulnerability} disentangles cross-modal instances into modality-related and modality-unrelated components, and uses the former to boost the reliability of the hashing network. 
\cite{li2020weakly} jointly studies the weakly-supervised semantic information and data structures for effective hashing retrieval.
\cite{jin2020deep} performs image-text and video-text retrieval via 2-D and 3-D CNNs respectively, in which both inter-modality and intra-modality information are considered.
To understand massive social images, \cite{li2018deeptpami} introduces a Deep Collaborative Embedding (DCE) network to learn common representations of images and tags.

\subsection{Uncertainty in Deep Learning} 
Here, uncertainty means the uncertainty of the deep neural network to the current outputs.
For traditional deep learning, the network only outputs a deterministic result.
However, in many scenarios, such as autonomous driving, we would like to simultaneously obtain the uncertainty of the network to that output. This will facilitate reliability assessment and risk-based decision \cite{chang2020data}.
Therefore, uncertainty has received much attention in recent years \cite{kendall2015bayesian,tang2020uncertainty}.
\cite{gal2016dropout} develops an approximate Bayesian inference framework to represent model uncertainty, which denotes the uncertainty existed in model parameters.
\cite{kendall2017uncertainties} proposes to estimate model uncertainty and data uncertainty (existed in the training data) in a unified framework.
\cite{wu2020unsupervised} utilizes uncertainty to learn a confidence map that facilitates 3D deformable modeling.
\cite{chang2020data} explores the uncertainty in face images with different qualities, significantly boosting recognition performance.
\cite{yu2019robust} introduces uncertainty into the feature learning of person re-identification. 

In summary, there are three common uncertainty estimation strategies. The first one estimates uncertainty based on the changes of the outputs of the network \cite{gal2016dropout,zheng2020rectifying}. For example, for the same input data, there will be multiple outputs by performing multiple different dropouts on the network during the inference phase. In this case, the variance of these outputs reflects the uncertainty. The second one integrates uncertainty into the objective function and directly outputs the uncertainty via the network \cite{kendall2017uncertainties,wu2020unsupervised}. The third one estimates uncertainty at the feature level \cite{chang2020data,yu2019robust}. For example, the feature is represented as a Gaussian distribution that is composed of learnable mean and variance, and the variance indicates the uncertainty. It is obvious that our method belongs to the first category.

\section{Preliminaries}\label{Preliminaries}
\subsection{Notation}
The notations employed in our method are listed in Table~\ref{table-notation}.
Concretely, uppercase letters such as $B$ are used to denote matrices, and $B_{ij}$ is used to denote the $(i, j)$-th element of $B$.
$B^{\top}$ indicates the transpose of the matrix $B$.
Lowercase letters like $b$ denote vectors.
$||b||_2$ is the Euclidean norm of the vector $b$.
$a^{\top}b$ denotes the product of vectors $a$ and $b$.
sign($\cdot$) means the element-wise sign function, which returns $+1$ and $-1$ when the element is positive and negative, respectively.

\begin{table}[t]
    \centering
    \caption{Meaning of the notations employed in our method.}
    \label{table-notation}
    \resizebox{0.45\textwidth}{!}{
    \begin{tabular}{|c||c|}
        \hline
        Notation & Meaning \\
        \hline
        $B$ & matrix \\
        \hline
        $B_{ij}$ & $(i, j)$-th element of matrix $B$ \\
        \hline
        $B^{\top}$ & transpose of matrix $B$ \\
        \hline
        $b$ & vector \\
        \hline
        $||b||_2$ & Euclidean norm of vector $b$ \\
        \hline
        $a^{\top}b$ & product of vectors $a$ and $b$ \\
        \hline
        sign($\cdot$) & element-wise sign function \\
        \hline
    \end{tabular}}
\end{table}

\subsection{Problem Definition}
Suppose there are a total of $n$ images $X = \{ x_i \}^n_{i=1}$, where $x_i$ means the $i$-th image.
For deep supervised hashing, the pairwise similarity between two images is also available.
The similarity matrix is denoted as $S$ with $S_{ij} \in \{ 0, 1 \}$, where $S_{ij} = 1$ means $x_i$ and $x_j$ are similar and $S_{ij} = 0$ means $x_i$ and $x_j$ are dissimilar.

The purpose of deep supervised hashing is to learn a function that maps the data from high-dimensional pixel space to low-dimensional binary Hamming space.
That is, for each image $x_i$, we can obtain a binary hashing code $b_i \in \{ -1, +1 \}^c$, where $c$ means that the hashing code has $c$ bits.
Meanwhile, the semantic similarity should be consistent before and after mapping.
For example, if $S_{ij}=1$, there should be as short Hamming distance as possible between $b_i$ and $b_j$.
Otherwise if $S_{ij}=0$, $b_i$ and $b_j$ should have long Hamming distance.
Hamming distance between two binary codes is defined as:
\begin{equation}\label{dis_H}
    dis_H(b_i, b_j) = \frac{1}{2}(c - b_i^{\top}b_j).
\end{equation}
As can be seen from the above definition, we need to find the optimal binary code for each data from all discrete $2^c$ possibilities, which is a classic combinatorial optimization problem.

\begin{figure}[t]
\centering
\includegraphics[width=0.5\textwidth]{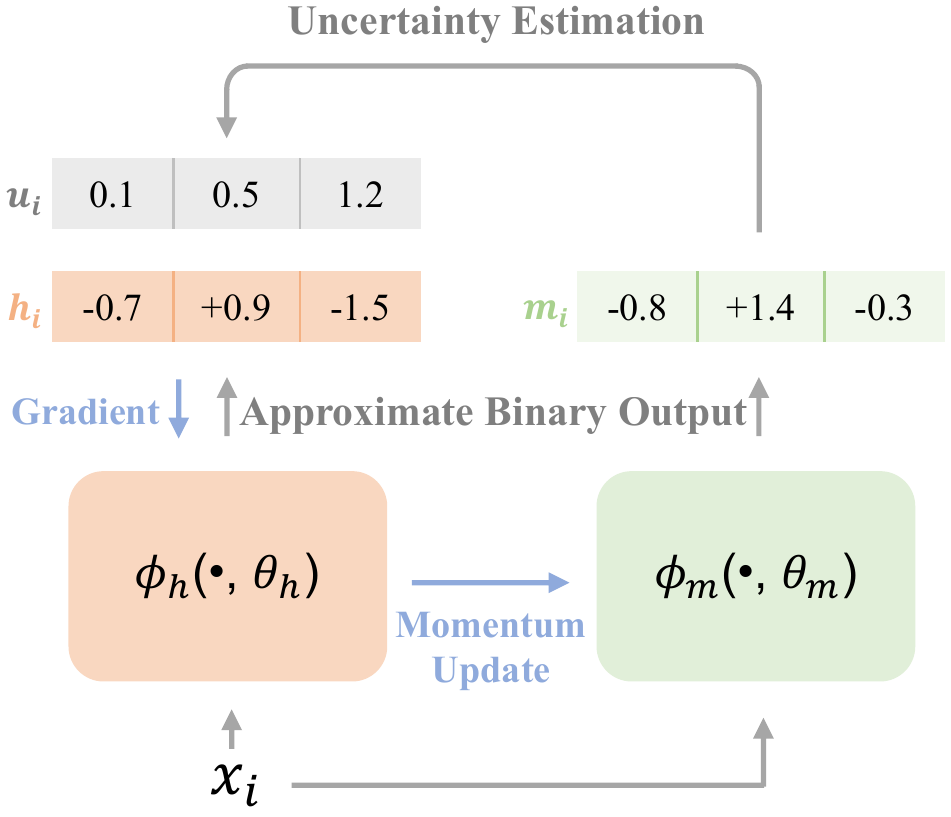}
\caption{Framework of our method that consists of a hashing network $\phi_{h}(\cdot, \theta_{h})$ and a momentum-updated network $\phi_{m}(\cdot, \theta_{m})$. The weights $\theta_{h}$ are updated via back-propagation of gradient, while the weights $\theta_{m}$ are updated by averaging $\theta_{h}$.
Given an input image $x_i$, besides outputting the approximate binary value $h_i = \phi_{h}(x_i, \theta_{h})$ as previous works, our method also outputs the uncertainty $u_i$. It derives from the discrepancy between $h_i$ and $m_i = \phi_{m}(x_i, \theta_{m})$.
}
\label{fig-framework}
\end{figure}

\section{Method}
In this section, we present the proposed DMUH in details, which integrates both bit-level and image-level uncertainty into the learning process of hashing codes. In the following parts of this section, we first introduce the overall framework of DMUH. Then, we revisit the traditional hashing learning algorithm and point out its potential problem. Subsequently, a novel uncertainty estimation approach is proposed. On this basis, we finally derive the uncertainty-aware hashing learning method and analyze its advantages over previous methods.

\subsection{Overall Framework}
As depicted in Fig.~\ref{fig-framework}, the framework of DMUH contains two networks: a hashing network $\phi_{h}(\cdot, \theta_{h})$ and a momentum-updated network $\phi_{m}(\cdot, \theta_{m})$, where $\theta_{h}$ and $\theta_{m}$ are the weights of the two networks, respectively.
Moreover, the two networks have a same architecture: a backbone for feature learning as well as a fully-connected layer for approximate binary coding.
The difference lies in that the weights $\theta_{h}$ are updated via back-propagation of gradient, while the weights $\theta_{m}$ are updated by averaging $\theta_{h}$.

Given an input image $x_i$, the two networks output approximate binary values $h_i = \phi_{h}(x_i, \theta_{h})$ and $m_i = \phi_{m}(x_i, \theta_{m})$, respectively.
Since $\phi_{m}(\cdot, \theta_{m})$ can be seen as an ensemble of $\phi_{h}(\cdot, \theta_{h})$ \cite{french2017self}, we can approximatively calculate the change of $h_i$ over previous values by comparing the discrepancy between $h_i$ and $m_i$.
As mentioned in Section~\ref{Introduction}, such change is regarded as the uncertainty of the hashing network to the current output value.
For each bit, the larger the difference between $h_i$ and $m_i$, the more uncertainty the hashing network to that bit. 
By this means, we can get the bit-level uncertainty.
For instance, as shown in Fig.~\ref{fig-framework}, it is obvious that the hashing network is more uncertain about the output of the third bit.
In addition, by averaging the uncertainty values of all bits in a hashing code, we can obtain the image-level uncertainty. 
It represents the uncertainty of the hashing network to the corresponding input image. 
After obtaining the bit-level and the image-level uncertainty, we will set different attention for different bits and images during training.
The detailed optimization process is introduced in the following parts.

\subsection{Hashing Learning Revisit}
Given the binary hashing codes $B$, the likelihood of the pairwise similarity $S$ is formulated as \cite{zhang2014supervised,li2015feature}:
\begin{equation}
p(S_{ij}|B) = \left\{
        \begin{array}{ll}
             \sigma(\Omega_{ij}), &s_{ij} = 1  \\
             1 - \sigma(\Omega_{ij}), &s_{ij} = 0 
        \end{array}
\right.
\end{equation}
where $\Omega_{ij} = \frac{1}{2} b_i^{\top}b_j$ and $\sigma(\Omega_{ij}) = \frac{1}{1 + e^{-\Omega_{ij}}}$.
Considering the negative log-likelihood of the pairwise similarity $S$, hashing codes $b_i$ and $b_j$ can be optimized by \cite{li2015feature}:
\begin{equation}\label{E_1}
    \mathcal{L} = -\log p(S|B) = - \sum_{S_{ij}} (S_{ij} \Omega_{ij} - \log(1 + e^{\Omega_{ij}})).
\end{equation}
Combining with Eq.~(\ref{dis_H}), we can find that minimizing Eq.~(\ref{E_1}) will make similar image pairs have shorter Hamming distance, and dissimilar image pairs have longer Hamming distance.
Given that the discrete binary hashing codes are not differentiable, a usual solution is to replace the discrete binary codes with continuous real-values. Subsequently, a regularization term is imposed to enforce the real-values to be close to binary ones \cite{li2015feature}:
\begin{equation}\label{E_2}
    \mathcal{L} = - \sum_{S_{ij}} (S_{ij} \Theta_{ij} - \log(1 + e^{\Theta_{ij}})) + \beta \sum_{i} ||h_i - b_i||_2^2,
\end{equation}
where $h_i = \phi_{h}(x_i, \theta_{h})$ are continuous real-values, $\Theta_{ij} = \frac{1}{2} h_i^{\top}h_j$, $b_i = $ sign$(h_i)$, and $\beta$ is a hyper-parameter.

\paragraph{Discussion}
Obviously, Eq.~(\ref{E_2}) can be used to learn hashing codes, where the first term optimizes the distance in Hamming space and the second term constrains the real-values to approximate binary codes.
However, Eq.~(\ref{E_2}) treats all hashing bits and input images equally, without considering their differences.
As shown in Fig.~\ref{fig-uncertainty}, the hashing network has different uncertainty to the hashing bits and the input images.
Therefore, we argue that each hashing bit and each input image should be treated separately according to the magnitude of the uncertainty, rather than being treated equally.

\subsection{Uncertainty Estimation}
During the training process of the hashing network, the output real-value of each hashing bit constantly changes to minimize the objective function Eq.~(\ref{E_2}). 
Intuitively, if the real-value of one bit changes a lot during the optimization, it indicates that the hashing network has high uncertainty to that bit. 
In order to measure this change, a straightforward approach is to store the output of each bit in each optimization step, and then compare the current output with the previous ones. However, it is unfeasible because of the requirement of huge storage memory when training on large-scale datasets. 

Inspired by the recently proposed momentum model in unsupervised and semi-supervised learning \cite{he2019momentum,tarvainen2017mean,french2017self}, we introduce a momentum-updated network $\phi_{m}(\cdot, \theta_{m})$ to help to estimate the uncertainty.
Different from the hashing network $\phi_{h}(\cdot, \theta_{h})$ that updates its weights $\theta_{h}$ via gradient back-propagation, $\phi_{m}(\cdot, \theta_{m})$ updates $\theta_{m}$ by averaging $\theta_{h}$:
\begin{equation}\label{E_3}
\theta_{m} = \alpha \theta_{m} + (1 - \alpha) \theta_{h},
\end{equation}
where $\alpha\in[0, 1)$ is a momentum coefficient hyper-parameter, whose value controls the smoothness of $\theta_{m}$. 
A larger $\alpha$ will result in smoother $\theta_{m}$.
Such an optimization manner can be seen as assembling the hashing networks in different optimization steps to the momentum-updated network \cite{french2017self}.
Therefore, comparing the output of the hashing network and that of the momentum-updated network, we can approximately obtain the change of each bit during training.
We regard this change as the bit-level uncertainty. 
That is, if a bit changes a lot, it means that the hashing network has high uncertainty to the current approximate value.
Formally, the uncertainty is defined as:
\begin{equation}\label{E_4}
    u_i = |h_i - m_i|,
\end{equation}
where $|\cdot|$ is an element-wise absolute value operation. $u_i$ is a vector and each element represents the bit-level uncertainty of the corresponding hashing bit. 
After getting the bit-level uncertainty, by counting the average uncertainty of all bits in a hashing code, we can further obtain the image-level uncertainty of the input image corresponding to that hashing code:
\begin{equation}\label{E_5}
    \bar{u_{i}} =  \frac{1}{c} \sum_{k=1}^{c}(u_i^k),
\end{equation}
where the image-level uncertainty $\bar{u_{i}}$ is a single value instead of a vector. 

\paragraph{Discussion} 
Fig.~\ref{fig-uncertainty} (Top) presents the approximate real-values of two bits at different training epochs. 
It is obvious that the two bits have different change trends. 
After 40 epochs, bit-1 still changes sharply, while bit-2 changes slightly.
In addition, the calculated uncertainty values through Eq.~(\ref{E_4}) of the two bits are 0.073 and 0.005, respectively.
We can see that the magnitude of the uncertainty is consistent with the change degree of the approximate real-value. 
That is, bit-1 has larger uncertainty and correspondingly has more drastic value change.
Therefore, it is reasonable to leverage the discrepancy between the output of the hashing network and that of the momentum-updated network to represent the uncertainty. 
Finally, Fig.~\ref{fig-uncertainty} (Bottom) displays the images with different image-level uncertainty.
We can find that the images with low uncertainty ($\bar{u_{i}} < 0.1$) usually have clear objects and single backgrounds, while the images with high uncertainty ($\bar{u_{i}} > 0.7$) contain nebulous objects and complex scenes.
For instance, it is difficult to recognize the frog from the first image in the bottom right corner.
These phenomena reveal the relationship between the image-level uncertainty and the input images.

\begin{algorithm}[t]
\caption{Optimization Algorithm}
\label{algorithm}
\textbf{Input}: \\
    Training set $X$, semantic similarity $S$ \\
\textbf{Output}: \\
    The weights of the hashing network $\theta_{h}$ and those of the momentum-updated network $\theta_{m}$ \\
\textbf{REPEAT}
\begin{algorithmic}
\STATE $\bullet$ Randomly sample a batch of training data with pairwise similarity;
\STATE $\bullet$ Compute the outputs of the hashing network and those of the momentum-updated network;
\STATE $\bullet$ Compute bit-level uncertainty and image-level uncertainty according to Eq.~(\ref{E_4}) and Eq.~(\ref{E_5}), respectively;
\STATE $\bullet$ Update $\theta_{h}$ according to Eq.~(\ref{E_8}) with standard gradient back-propagation;
\STATE $\bullet$ Update $\theta_{m}$ according to Eq.~(\ref{E_3});
\end{algorithmic}
\textbf{UNTIL} a fixed number of iterations
\end{algorithm}

\subsection{Uncertainty-aware Hashing Learning}
After getting the bit-level uncertainty $u_i$, we leverage it to guide the optimization of the regularization.
Rather than treating each bit equally as Eq.~(\ref{E_2}), we set different weights for different bits according to the magnitude of the uncertainty, yielding a new optimization objective:
\begin{equation}\label{E_6}
    \mathcal{L} = - \sum_{S_{ij}} (S_{ij} \Theta_{ij} - \log(1 + e^{\Theta_{ij}})) + \beta \sum_{i} e^{u_i} ||h_i - b_i||_2^2,
\end{equation}
where $e^{u_i}$ is multiplied as a weight on the regularization term. The hashing bit with higher uncertainty is given a larger weight during regularization.
In addition, the image-level uncertainty allows us to set different weights for different input images.
We apply larger weights to the images with higher uncertainty in the optimization of Hamming distance.
Considering both the uncertainty $\bar{u_{i}}$ and $\bar{u_{j}}$ of images $x_i$ and $x_j$, Eq.~(\ref{E_6}) is reformulated as:
\begin{equation}\label{E_7}
    \mathcal{L} = - \sum_{S_{ij}} e^{\bar{u_i}+\bar{u_j}} (S_{ij} \Theta_{ij} - \log(1 + e^{\Theta_{ij}})) + \beta \sum_{i} e^{u_i} ||h_i - b_i||_2^2.
\end{equation}
It is obvious that Eq.~(\ref{E_7}) separately treats different input images (the first term) and hashing bits (the second term) under the guidance of the image-level uncertainty and the bit-level uncertainty, respectively.
On this basis, we further involve the uncertainty into the optimization objective:
\begin{equation}\label{E_8}
    \begin{aligned}
    \mathcal{L} = & - \sum_{S_{ij}} e^{\bar{u_i}+\bar{u_j}} (S_{ij} \Theta_{ij} - \log(1 + e^{\Theta_{ij}})) \\
    & + \beta \sum_{i} e^{u_i} ||h_i - b_i||_2^2 + \gamma \sum_{i} u_i,
    \end{aligned}
\end{equation}
where $\gamma$ is a trade-off parameter.
The whole optimization process for the hashing network and the momentum-updated network is summarized in Algorithm~\ref{algorithm}.

\paragraph{Discussion}
\emph{What are the advantages of the uncertainty-aware hashing learning?}
To begin with, according to the observations in Fig.~\ref{fig-uncertainty}, hard examples can be discovered automatically based on the magnitude of the image-level uncertainty. 
Benefiting from this, the first term of Eq.~(\ref{E_8}) can focus on the optimization of the hard examples.
The effectiveness of such a hard example based optimization has been fully proved in previous works \cite{lin2017focal,wu2017sampling}.
Furthermore, the second term of Eq.~(\ref{E_8}) assists in stabilizing the outputs of the bits that change frequently, which may accelerate the convergence of the hashing network.
Finally, the third term of Eq.~(\ref{E_8}) minimizes the discrepancy between the outputs of the hashing network and those of the momentum-updated network.
Since the momentum-updated network is actually an ensemble of the hashing networks in different optimization steps, the third term of Eq.~(\ref{E_8}) will help to improve the retrieval performance of the hashing network \cite{french2017self}.

\section{Experiments}
In this section, we systematically analyze the proposed DMUH and compare it against state-of-the-art methods on four popular datasets, including CIFAR-10 \cite{krizhevsky2009learning}, Clothing1M \cite{xiao2015learning}, NUS-WIDE \cite{chua2009nus}, and MS-COCO \cite{lin2014microsoft}. The remainder of this section is organized as follows. We start with introducing the used datasets and the corresponding protocols. Then, experimental details of our method are reported. Subsequently, insightful experimental analyses of our method are provided. Finally, comprehensive comparisons with state-of-the-art methods are given. 

\subsection{Datasets and Protocols}
A total of four widely used datasets are employed to evaluate the proposed method, including two single-label (each image merely belongs to one class) datasets CIFAR-10 and Clothing1M, as well as two multi-label (each image belongs to one or multiple classes) datasets NUS-WIDE and MS-COCO. 

\paragraph{CIFAR-10} It consists of 60,000 color images in 32$\times$32 resolution from 10 classes with 6,000 images per class. The labeled 10 classes include 'airplane', 'automobile', 'bird', 'cat', 'deer', 'dog', 'frog', 'horse', 'ship', and 'truck'. 
Following the protocol in \cite{li2015feature}, we randomly sample 1,000 images with 100 images per class as the query set, and randomly select 5,000 images with 500 images per class from the rest images as the training set. Other images are used as the database set.

\paragraph{Clothing1M} It is a million-level large-scale dataset with a total of 1,037,497 images that are collected from the online shopping website. The clarified classes are various clothes, such as 'jacket', 't-shirt', 'shawl', 'downcoat', 'hoodie', and 'sweater'.
Following the protocol in \cite{jiang2018deep}, 7,000 images are randomly sampled as the query set and 14,000 images are randomly selected as the training set. About a million images are utilized as the database set.

\paragraph{NUS-WIDE} It contains 269,648 images collected from the Flickr website. Each image is annotated with one or multiple labels from 81 classes, including 'water', 'clouds', 'ocean', 'road', 'buildings', 'toy', 'window', 'zebra', 'sun', 'street', and so on. Following the protocol in \cite{li2015feature}, only 195,834 images belonging to the 21 most frequent classes are leveraged in our experiments. 2,100 images with 100 images per class are randomly sampled as the query set and 10,500 images with 500 images per class are randomly selected as the training set. Other images are leveraged as the database set.

\paragraph{MS-COCO} It has 82,783 training images and 40,504 validation images that are collected from the website. Each image belongs to one or multiple labels from 91 classes, including 'car', 'cat', 'plate', 'oven', 'pizza', 'clock', 'bird', 'boat', 'airplane', 'cake', 'laptop', 'book', 'cup', 'suitcase', 'apple', and so on.
Following the protocol in \cite{jiang2017asymmetric}, 5,000 images are randomly sampled as the query set and 10,000 images are randomly selected as the training set. Other images are set as the database set.

\begin{table}[t]
    \centering
    \caption{Architecture of the hashing network and the momentum-updated network. K/S/P denotes kernel size/stride/padding.}
    \label{table-network}
    \resizebox{0.35 \textwidth}{!}{
        \begin{tabular}{|ccc|}
            \hline
            LAYER & K/S/P/Pool & OUTPUT  \\
            \hline
            conv1 & 11/4/0/2 & 64$\times$27$\times$27  \\
            conv2 & 5/1/2/2 & 256$\times$13$\times$13  \\
            conv3 & 3/1/1/0 & 256$\times$13$\times$13  \\
            conv4 & 3/1/1/0 & 256$\times$13$\times$13  \\
            conv5 & 3/1/1/2 & 256$\times$6$\times$6  \\
            \hline
            full6 & - & 4096 \\
            full7 & - & 4096 \\
            full8 & - & code length \\
            \hline
        \end{tabular}}
\end{table}

\paragraph{Evaluation Methodology}
Following the setting of \cite{lai2015simultaneous}, Mean Average Precision (MAP) is adopted to evaluate retrieval performance. 
Concretely, given a query image $x_q$, average precision (AP) is defined as \cite{jiang2018deep}:
\begin{equation}
    AP(x_q) = \frac{1}{R_k} \sum_{k} P(k) \mathds{I}_1(k),
\end{equation}
where $R_k$ denotes the number of all relevant images. $P(k)$ denotes the precision at the cut-off $k$ in the returned image list after retrieval.
$\mathds{I}_1(k)$ is an indicator function, which is equal to $1$ if the $k$-th returned images is similar with $x_q$ and is equal to $0$ when the $k$-th returned images is dissimilar with $x_q$. 
MAP is the mean AP of all $Q = \{1,...,q\}$ queries:
\begin{equation}
    MAP = \frac{1}{Q} \sum_{q} AP(x_q).
\end{equation}
Particularly, for the NUS-WIDE dataset, the MAP is calculated within the top 5,000 returned neighbors. For the single-label CIFAR-10 and Clothing1M datasets, two images are treated as a similar pair ($S_{ij} = 1$) when they come from a same class, otherwise they are considered as a dissimilar pair. For the multi-label NUS-WIDE and MS-COCO datasets, two images are regarded as a similar pair if they share at least one common label.

\subsection{Experimental Details}
For a fair comparison with other state-of-the-art methods, we employ the CNN-F network \cite{chatfield2014return} pre-trained on ImageNet as the backbone of the hashing network.
As shown in Table~\ref{table-network}, the CNN-F network consists of five convolutional layers and three fully connected layers, where the last fully connected layer is modified as the hashing layer with the length of hashing codes (12 bits, 24 bits, 32 bits, and 48 bits).
The momentum-updated network has the same architecture as the hashing network.
The input images are first resized to 256$\times$256 resolution and then cropped to 224$\times$224.
Stochastic Gradient Descent (SGD) is used as the optimizer with 1e-4 weight decay.
The initial learning rate is set to 0.05 and gradually reduced to 0.0005.
The batch size is set to 128.
The momentum coefficient hyper-parameter $\alpha$ in Eq.~(\ref{E_3}) is set to 0.7, and the hyper-parameters $\beta$ and $\gamma$ in Eq.~(\ref{E_8}) are set to 50 and 1, respectively.
We determine the values of $\beta$ and $\gamma$ by balancing the magnitude of the corresponding loss term.
All experiments are conducted on a single NVIDIA TITAN RTX GPU.

\begin{figure}[t]
\centering
\includegraphics[width=0.8\textwidth]{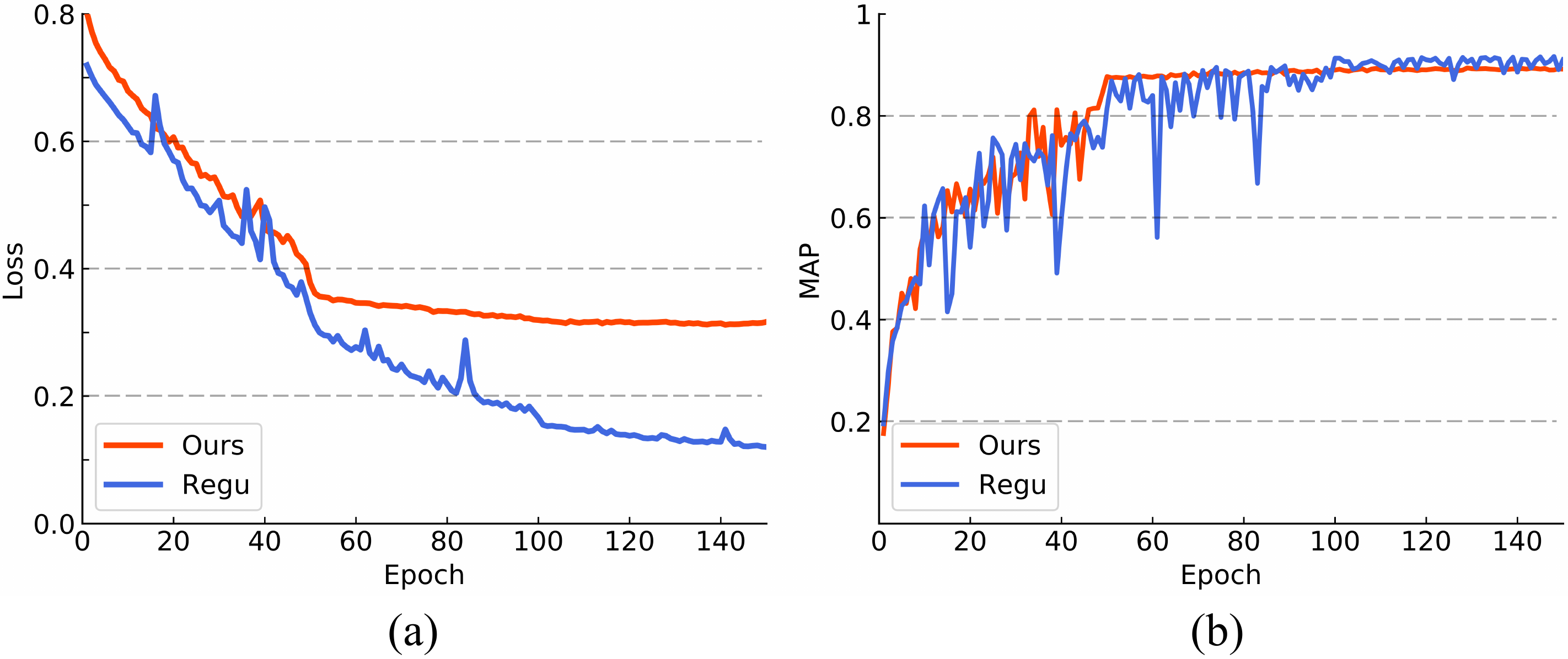}
\caption{Loss and MAP curves during training on the CIFAR-10 dataset. Furthermore, the MAP values of our method and \emph{Regu} in the testing phase are 0.815 and 0.739, respectively.}
\label{fig-loss_map}
\end{figure}

\subsection{Evaluation of the Uncertainty-aware Hashing}
In this subsection, we compare our proposed method against the traditional regularization based method (denoted as \emph{Regu}), whose optimization objective is Eq.~(\ref{E_2}).
The only difference between our method and \emph{Regu} is the introduced uncertainty, including its estimation and usage.

Fig.~\ref{fig-loss_map} (a) depicts the loss curves of the two methods under different training epochs on the CIFAR-10 dataset.
Fig.~\ref{fig-loss_map} (b) plots the corresponding MAP curves on the training set.
Observing the results, we can see that the loss curve of our method converges after about 50 epochs.
Meanwhile, the MAP curve reaches its peak and remains stable.
By contrast, the loss curve of \emph{Regu} converges much slower. 
In particular, the MAP curve of \emph{Regu} still sharply oscillates at the 90-th epoch.
The faster convergence of our method may be due to that we pay more attention on the hashing bits with drastic value changes (benefiting from the bit-level uncertainty) and the hard examples (benefiting from the image-level uncertainty).
In addition, although the two methods finally have a similar MAP value in the training phase, our method obtains a much better MAP result (0.815) than \emph{Regu} (0.739) in the testing phase.
This further demonstrates the great generalization ability of our uncertainty based method.

\begin{figure*}[t]
\centering
\includegraphics[width=0.985\textwidth]{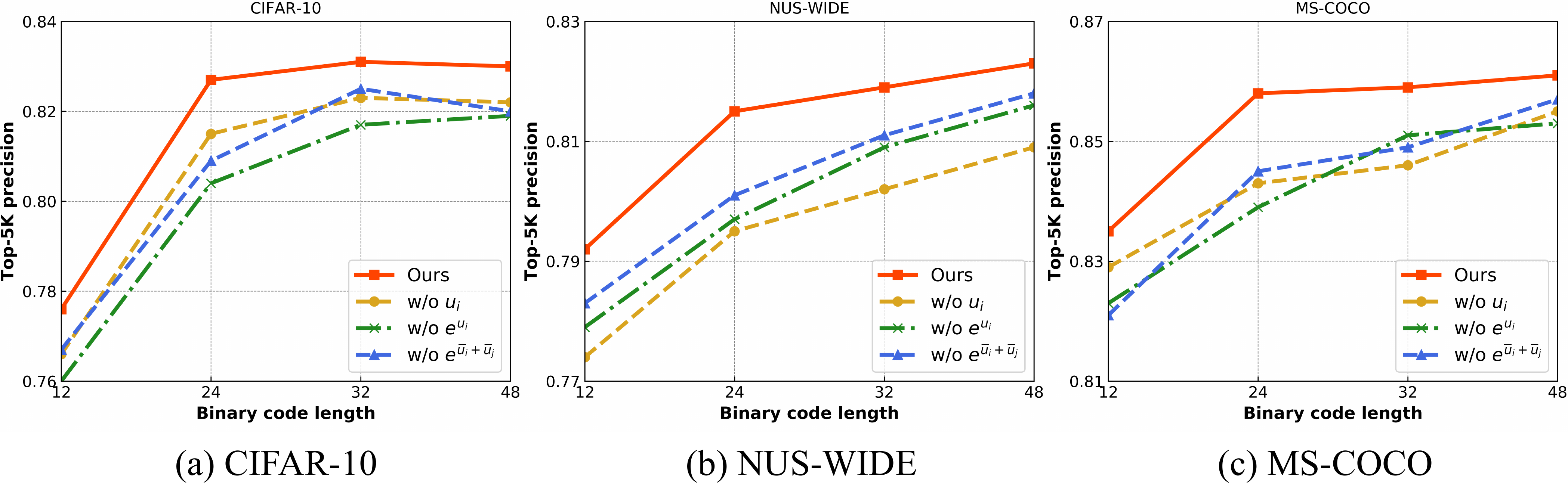}
\caption{Top-5K precisions on the CIFAR-10, NUS-WIDE, and MS-COCO datasets.}
\label{fig-top-k_precision}
\end{figure*}

\subsection{Ablation Study}
In this subsection, we compare our method against its three variants to reveal the role of each component.
Among them, $w/o~u_i$ means removing the optimization of uncertainty, \textit{i.e.} the third term of Eq.~(\ref{E_8}).
In such a case, we no longer minimize the output discrepancy between the hashing network and the momentum-updated network.
$w/o~e^{u_i}$ denotes discarding the bit-level uncertainty $e^{u_i}$ in the second term of Eq.~(\ref{E_8}), which means that we ignore the differences among hashing bits.
$w/o~e^{\bar{u_i}+\bar{u_j}}$ represents removing the image-level uncertainty in the first term of Eq.~(\ref{E_8}).
At this point, all training images are treated equally during training.

Following the setting of \cite{jiang2017asymmetric}, we report Top-5K precision curves to measure retrieval performance on the CIFAR-10, NUS-WIDE, and MS-COCO datasets.
The comparison results are reported in Fig.~\ref{fig-top-k_precision}.
It is observed that our method obtains the best retrieval performance.
The improvements of our method over $w/o~u_i$ suggest the impact of the uncertainty minimizing, which assists in transferring the knowledge from the momentum-updated network to the hashing network \cite{tarvainen2017mean}.
The gains of our method over $w/o~e^{u_i}$ demonstrate the effectiveness of the bit-level uncertainty.
The hashing bits with drastic value changes are given larger weights to stabilize their outputs.
The improvements of our method over $w/o~e^{\bar{u_i}+\bar{u_j}}$ prove the validity of the image-level uncertainty. It enables the hard examples to receive more attention during optimization and thereby helps to improve retrieval performance.

Furthermore, we also give detailed parameter analyses.
Table~\ref{table-2} reports the parameter study of the trade-off parameters $\beta$ and $\gamma$ in Eq.~(\ref{E_8}), and the momentum coefficient $\alpha$ in Eq.~(\ref{E_3}).
As can be seen from Tables~\ref{table-2} (a) and (b), our method is not sensitive to $\beta$ and $\gamma$ in a large range. 
For example, the MAP value of 24 bits only changes 0.007 when $\beta$ is set from 30 to 70.
Table~\ref{table-2} (c) suggests that the optimal value of $\alpha$ is 0.7.

\begin{table}[t]
    \centering
    \caption{Parameter analyses on CIFAR-10 dataset, including the trade-off parameters $\beta$ and $\gamma$ in Eq.~(\ref{E_8}), and the momentum coefficient $\alpha$ in Eq.~(\ref{E_3}).}
    \label{table-2}
    \subfloat[$\beta$ in Eq.~(\ref{E_8})]{
        \label{eta}
        \resizebox{!}{0.043\textheight}{
        \begin{tabular}{c|c|c|c|c}
                \toprule[1pt]
                $\beta$ & 12 bits & 24 bits & 32 bits & 48 bits \\
                \hline
                30 & 0.764 & 0.808 & 0.815 & 0.819 \\
                40 & 0.769 & 0.810 & 0.818 & 0.822 \\
                50 & 0.772 & 0.815 & 0.822 & 0.826 \\
                60 & 0.767 & 0.811 & 0.819 & 0.820 \\
                70 & 0.765 & 0.809 & 0.817 & 0.816 \\
                \bottomrule[1pt]
        \end{tabular}}
    }
    \subfloat[$\gamma$ in Eq.~(\ref{E_8})]{
        \label{alpha}
        \resizebox{!}{0.043\textheight}{
        \begin{tabular}{c|c|c|c|c}
                \toprule[1pt]
                $\gamma$ & 12 bits & 24 bits & 32 bits & 48 bits \\
                \hline
                0.2 & 0.760 & 0.806 & 0.817 & 0.817 \\
                0.5 & 0.765 & 0.810 & 0.819 & 0.821 \\
                1   & 0.772 & 0.815 & 0.822 & 0.826 \\
                2   & 0.769 & 0.809 & 0.814 & 0.823 \\
                3   & 0.768 & 0.807 & 0.810 & 0.819 \\
                \bottomrule[1pt]
        \end{tabular}}
    }
    \subfloat[$\alpha$ in Eq.~(\ref{E_3})]{
        \label{interval}
        \resizebox{!}{0.043\textheight}{
        \begin{tabular}{c|c|c|c|c}
                \toprule[1pt]
                $\alpha$  & 12 bits & 24 bits & 32 bits & 48 bits \\
                \hline
                0.5 & 0.766 & 0.807 & 0.813 & 0.821 \\
                0.6 & 0.768 & 0.810 & 0.817 & 0.822 \\
                0.7 & 0.772 & 0.815 & 0.822 & 0.826 \\
                0.8 & 0.772 & 0.813 & 0.819 & 0.825 \\
                0.9 & 0.769 & 0.810 & 0.815 & 0.819 \\
                \bottomrule[1pt]
        \end{tabular}}
    }
\end{table}

\begin{table*}[t]
    \begin{center}
        \caption{MAP of different methods on the single-label datasets CIFAR-10 and Clothing1M.}
        \label{table-1}
        \resizebox{0.94 \textwidth}{!}{
        \begin{tabular}{l|cccc|cccc}
            \toprule[1.1pt]
            \multirow{2}{*}{Method}
            & \multicolumn{4}{|c}{CIFAR-10}
            & \multicolumn{4}{|c}{Clothing1M} \\
            \cline{2-5}
            \cline{6-9}
                    & 12 bits & 24 bits & 32 bits & 48 bits & 12 bits & 24 bits & 32 bits & 48 bits \\
            \hline
            \textbf{DMUH}    &\bf 0.772  &\bf 0.815  & \bf 0.822 & \bf 0.826 & \bf 0.315 & \bf 0.371 & \bf 0.389 & \bf 0.401 \\
            \hline
            \hline
            DDSH    & 0.753     & 0.776     & 0.803     & 0.811     & 0.271 & 0.332 & 0.343 & 0.346 \\
            DSDH    & 0.740     & 0.774     & 0.792     & 0.813     & 0.278 & 0.302 & 0.311 & 0.319 \\
            DPSH    & 0.712     & 0.725     & 0.742     & 0.752     & 0.193 & 0.204 & 0.213 & 0.215 \\
            DSH     & 0.644     & 0.742     & 0.770     & 0.799     & 0.173 & 0.187 & 0.191 & 0.202 \\
            DHN     & 0.680     & 0.721     & 0.723     & 0.733     & 0.190 & 0.224 & 0.212 & 0.248 \\
            \hline
            COSDISH & 0.583     & 0.661     & 0.680     & 0.701     & 0.187 & 0.235 & 0.256 & 0.275 \\
            SDH     & 0.453     & 0.633     & 0.651     & 0.660     & 0.151 & 0.186 & 0.194 & 0.197 \\
            FastH   & 0.597     & 0.663     & 0.684     & 0.702     & 0.173 & 0.206 & 0.216 & 0.244 \\
            LFH     & 0.417     & 0.573     & 0.641     & 0.692     & 0.154 & 0.159 & 0.212 & 0.257 \\
            \hline
            ITQ     & 0.261     & 0.275     & 0.286     & 0.294     & 0.115 & 0.121 & 0.122 & 0.125 \\
            \bottomrule[1.1pt]
        \end{tabular}}
    \end{center}
\end{table*}

\subsection{Comparisons with State-of-the-Art Methods}
In this subsection, the proposed method is evaluated against a total of 11 state-of-the-art hashing methods, including iterative quantization (ITQ) \cite{gong2011iterative}, column sampling based discrete supervised hashing (COSDISH) \cite{kang2016column}, supervised discrete hashing (SDH) \cite{shen2015supervised}, fast supervised hashing (FastH) \cite{lin2014fast}, latent factor hashing (LFH) \cite{zhang2014supervised}, deep supervised discrete hashing (DSDH) \cite{li2020general}, deep discrete supervised hashing (DDSH) \cite{jiang2018deep}, deep pairwise-supervised hashing (DPSH) \cite{li2015feature}, deep supervised hashing (DSH) \cite{liu2016deep}, deep hashing network (DHN) \cite{zhu2016deep}, and central similarity quantization (CSQ) \cite{yuan2020central}.
The brief introductions of these methods are as follows:
\begin{itemize}
  \item[-] ITQ aims to search a rotation of zero-centered data to bridge the quantization discrepancy.
  \item[-] COSDISH iteratively samples columns from the similarity matrix and hashing codes are alternatively optimized without relaxation.
  \item[-] SDH learns hashing codes for linear classification, which is solved by discrete cyclic coordinate descent.
  \item[-] FastH introduces decision trees as hashing coding functions, where the decision trees are learned by a correlative two-step approach.
  \item[-] LFH proposes to leverage latent factor models to learn similarity-preserving binary hashing codes. 
  \item[-] DSDH takes advantages of both similarity and classification information to learn hashing codes in a one-stream framework.
  \item[-] DDSH enhances the feedback between hashing coding and deep feature learning via a discrete optimization algorithm.
  \item[-] DPSH performs joint learning of hashing codes and deep features in an end-to-end framework.
  \item[-] DSH relaxes binary hashing codes to be real-values and adopts a pairwise training strategy to optimize Hamming distance.
  \item[-] DHN employs both a pairwise cross-entropy loss and a pairwise quantization loss to improve hashing quality.
  \item[-] CSQ presents a global central similarity and encourages the hashing codes of similar images to arrive at the corresponding centers. 
\end{itemize}

\begin{table*}[t]
    \begin{center}
        \caption{MAP of different methods on the multi-label datasets NUS-WIDE and MS-COCO. For the NUS-WIDE dataset, the MAP is calculated within the top 5,000 returned neighbors.}
        \label{table-3}
        \resizebox{0.94 \textwidth}{!}{
        \begin{tabular}{l|cccc|cccc}
            \toprule[1.1pt]
            \multirow{2}{*}{Method}
            & \multicolumn{4}{|c}{NUS-WIDE}
            & \multicolumn{4}{|c}{MS-COCO} \\
            \cline{2-5}
            \cline{6-9}
                    & 12 bits & 24 bits & 32 bits & 48 bits & 12 bits & 24 bits & 32 bits & 48 bits \\
            \hline
            \textbf{DMUH}    & \bf 0.792 & \bf 0.818 & \bf 0.825 & \bf 0.829 & \bf 0.761   & \bf 0.779  & \bf 0.785 & \bf 0.788 \\
            \hline
            \hline
            DDSH    & 0.776     & 0.803     & 0.810     & 0.817     & 0.745 & 0.765 & 0.771 & 0.774 \\
            DSDH    & 0.774     & 0.801     & 0.813     & 0.819     & 0.743 & 0.762 & 0.765 & 0.769 \\
            DPSH    & 0.768     & 0.793     & 0.807     & 0.812     & 0.741 & 0.759 & 0.763 & 0.771 \\
            DSH     & 0.712     & 0.731     & 0.740     & 0.748     & 0.696 & 0.717 & 0.715 & 0.722 \\
            DHN     & 0.771     & 0.801     & 0.805     & 0.814     & 0.744 & 0.765 & 0.769 & 0.774 \\
            \hline
            COSDISH & 0.642     & 0.740     & 0.784     & 0.796     & 0.689 & 0.692 & 0.731 & 0.758 \\
            SDH     & 0.764     & 0.799     & 0.801     & 0.812     & 0.695 & 0.707 & 0.711 & 0.716 \\
            FastH   & 0.726     & 0.769     & 0.781     & 0.803     & 0.719 & 0.747 & 0.754 & 0.760 \\
            LFH     & 0.711     & 0.768     & 0.794     & 0.813     & 0.708 & 0.738 & 0.758 & 0.772 \\
            \hline
            ITQ     & 0.714     & 0.736     & 0.745     & 0.755     & 0.633 & 0.632 & 0.630 & 0.633 \\
            \bottomrule[1.1pt]
        \end{tabular}}
    \end{center}
\end{table*}

The above state-of-the-art methods consist of three types of hashing learning approaches.
ITQ is a representative unsupervised learning method. 
COSDISH, SDH, FastH, and LFH are non-deep supervised learning methods. 
DSDH, DDSH, DPSH, DSH, DHN, and CSQ are deep supervised learning methods.

The comparison results of our method against the state-of-the-art methods are tabulated in Table~\ref{table-1}, Table~\ref{table-3}, and Table~\ref{table-new}, from which we obtain three observations.
First, the unsupervised method ITQ lags behind all of the supervised methods, suggesting the great advantage of the supervised information.
Second, the performance of deep supervised hashing methods is generally better than that of the non-deep supervised hashing methods.
This indicates that the features extracted by deep neural networks are better than the hand-crafted features.
Third, our method obtains the highest retrieval accuracy on all of the four datasets.
For example, on the CIFAR-10 dataset, our method surpasses DDSH by 3.9\% at 24 bits.
On the NUS-WIDE dataset, we improve the best MAP values of all bits at least 1.2\%.
On the MS-COCO dataset, we also get an improvement of 1.6\% at 12 bits.
Specially, on the large-scale Clothing1M dataset, the MAP value is significantly improved by \textbf{3.7\%}, \textbf{3.9\%}, \textbf{4.6\%}, and \textbf{5.5\%} in terms of 12, 24, 32, and 48 bits, respectively.
Compared with the state-of-the-art CSQ, our method also presents more superior performance, obtaining at least 1.8\% improvements on the four datasets.
The compared deep supervised hashing methods adopt a similar binary approximation that treats all hashing bits equally, such as DSDH uses the activation function \emph{Tanh} and DPSH uses the regularization.
With these in mind, we owe the gains of our method over the competitors to the proposed uncertainty-aware learning approach. It applies different attention weights for different hashing bits and input images according to the magnitude of the bit-level and image-level uncertainty, respectively.

\begin{table*}[t]
    \begin{center}
        \caption{Comparisons with CSQ. Since the code length of CSQ can only be 2$^n$, where $n \in$ (1, 2, 3, ...), we conduct experiments under 16 bits, 32 bits, and 64 bits that are consistent with the settings in \cite{yuan2020central}.}
        \label{table-new}
        \resizebox{0.96 \textwidth}{!}{
        \begin{tabular}{l|ccc|ccc|ccc|ccc}
            \toprule[1.1pt]
            \multirow{2}{*}{Method}
            & \multicolumn{3}{|c}{CIFAR-10}
            & \multicolumn{3}{|c}{Clothing1M} 
            & \multicolumn{3}{|c}{NUS-WIDE} 
            & \multicolumn{3}{|c}{MS-COCO} \\
            \cline{2-4}
            \cline{5-7}
            \cline{8-10}
            \cline{11-13}
                    & 16 bits & 32 bits & 64 bits & 16 bits & 32 bits & 64 bits 
                    & 16 bits & 32 bits & 64 bits & 16 bits & 32 bits & 64 bits \\
            \hline
            DMUH             & 0.779 & 0.822 & 0.830 & 0.320 & 0.389 & 0.402 
                             & 0.803 & 0.825 & 0.834 & 0.765 & 0.785 & 0.792 \\
            \hline
            CSQ              & 0.501 & 0.533 & 0.572 & 0.302 & 0.308 & 0.317
                             & 0.755 & 0.783 & 0.791 & 0.670 & 0.681 & 0.707 \\
            \bottomrule[1.1pt]
        \end{tabular}}
    \end{center}
\end{table*}

\section{Conclusion}
In this paper, we have proposed an uncertainty-aware deep supervised hashing method that is named as DMUH.
To begin with, we discover that the hashing network has different uncertainty to different approximate hashing bits.
According to this observation, we propose that hashing bits should be paid separate attention during training, rather than being treated equally.
Subsequently, we introduce a momentum-updated network to assist in estimating such bit-level uncertainty.
In addition, the mean bit-level uncertainty of all bits in a hashing code is seen as image-level uncertainty, which reflects the uncertainty of the hashing network to the corresponding input image.
The bit-level uncertainty and image-level uncertainty are leveraged to guide the regularization of hashing bits and facilitate the optimization of Hamming distance, respectively.
Finally, extensive experiments on the CIFAR-10, Clothing1M, NUS-WIDE, and MS-COCO datasets demonstrate the superiority of our proposed method over state-of-the-art counterparts, especially on the million-scale Clothing1M dataset.

In general, as far as we know, we are the first to study the uncertainty in binary bits, which can bring some useful insights to deep hashing methods and other similar discrete optimization problems.
However, there is still room for improvement in the proposed method, which optimizes approximate continued values of binary bits rather than directly optimizing binary values.
In the future, we will focus more on the discrete coding procedure.
Moreover, we also plan to extend our method to cross-modal hashing due to its wide application prospects.

\bibliography{mybibfile}

\begin{thebibliography}{10}
\expandafter\ifx\csname url\endcsname\relax
  \def\url#1{\texttt{#1}}\fi
\expandafter\ifx\csname urlprefix\endcsname\relax\def\urlprefix{URL }\fi
\expandafter\ifx\csname href\endcsname\relax
  \def\href#1#2{#2} \def\path#1{#1}\fi

\bibitem{bengio2020machine}
Y.~Bengio, A.~Lodi, A.~Prouvost, Machine learning for combinatorial
  optimization: a methodological tour d’horizon, European Journal of
  Operational Research.

\bibitem{yan2015multi}
J.~Yan, M.~Cho, H.~Zha, X.~Yang, S.~M. Chu, Multi-graph matching via affinity
  optimization with graduated consistency regularization, IEEE Transactions on
  Pattern Analysis and Machine Intelligence 38~(6) (2015) 1228--1242.

\bibitem{vinyals2015pointer}
O.~Vinyals, M.~Fortunato, N.~Jaitly, Pointer networks, in: Advances in Neural
  Information Processing Systems, 2015.

\bibitem{wang2020combinatorial}
R.~Wang, J.~Yan, X.~Yang, Combinatorial learning of robust deep graph matching:
  an embedding based approach, IEEE Transactions on Pattern Analysis and
  Machine Intelligence.

\bibitem{wang2018survey}
J.~Wang, T.~Zhang, J.~Song, N.~Sebe, H.~T. Shen, A survey on learning to hash,
  IEEE Transactions on Pattern Analysis and Machine Intelligence 40~(4) (2017)
  769--790.

\bibitem{lin2014fast}
G.~Lin, C.~Shen, Q.~Shi, A.~Van~den Hengel, D.~Suter, Fast supervised hashing
  with decision trees for high-dimensional data, in: IEEE Conference on
  Computer Vision and Pattern Recognition, 2014.

\bibitem{he2014robust}
R.~He, B.~Hu, X.~Yuan, L.~Wang, Robust recognition via information theoretic
  learning, 2014.

\bibitem{he2009robust}
R.~He, B.-G. Hu, X.-T. Yuan, Robust discriminant analysis based on
  nonparametric maximum entropy, in: Asian Conference on Machine Learning,
  2009.

\bibitem{gong2011iterative}
Y.~Gong, S.~Lazebnik, Iterative quantization: A procrustean approach to
  learning binary codes, in: IEEE Conference on Computer Vision and Pattern
  Recognition, 2011.

\bibitem{gionis1999similarity}
A.~Gionis, P.~Indyk, R.~Motwani, Similarity search in high dimensions via
  hashing, in: Very Large Data Bases, 1999.

\bibitem{li2015feature}
W.-J. Li, S.~Wang, W.-C. Kang, Feature learning based deep supervised hashing
  with pairwise labels, in: International Joint Conference on Artificial
  Intelligence, 2016.

\bibitem{fu2018neurons}
C.~Fu, L.~Song, X.~Wu, G.~Wang, R.~He, Neurons merging layer: towards
  progressive redundancy reduction for deep supervised hashing, in:
  International Joint Conference on Artificial Intelligence, 2019.

\bibitem{xia2014supervised}
R.~Xia, Y.~Pan, H.~Lai, C.~Liu, S.~Yan, Supervised hashing for image retrieval
  via image representation learning, in: AAAI Conference on Artificial
  Intelligence, 2014.

\bibitem{jiang2018deep}
Q.-Y. Jiang, X.~Cui, W.-J. Li, Deep discrete supervised hashing, IEEE
  Transactions on Image Processing 27~(12) (2018) 5996--6009.

\bibitem{liu2016deep}
H.~Liu, R.~Wang, S.~Shan, X.~Chen, Deep supervised hashing for fast image
  retrieval, in: IEEE Conference on Computer Vision and Pattern Recognition,
  2016.

\bibitem{lin2017focal}
T.-Y. Lin, P.~Goyal, R.~Girshick, K.~He, P.~Doll{\'a}r, Focal loss for dense
  object detection, in: IEEE International Conference on Computer Vision, 2017.

\bibitem{wu2017sampling}
C.-Y. Wu, R.~Manmatha, A.~J. Smola, P.~Krahenbuhl, Sampling matters in deep
  embedding learning, in: IEEE International Conference on Computer Vision,
  2017.

\bibitem{he2019momentum}
K.~He, H.~Fan, Y.~Wu, S.~Xie, R.~Girshick, Momentum contrast for unsupervised
  visual representation learning, in: IEEE Conference on Computer Vision and
  Pattern Recognition, 2020.

\bibitem{tarvainen2017mean}
A.~Tarvainen, H.~Valpola, Mean teachers are better role models: Weight-averaged
  consistency targets improve semi-supervised deep learning results, in:
  Advances in Neural Information Processing Systems, 2017.

\bibitem{french2017self}
G.~French, M.~Mackiewicz, M.~Fisher, Self-ensembling for visual domain
  adaptation, in: International Conference on Learning Representations, 2018.

\bibitem{krizhevsky2009learning}
A.~Krizhevsky, G.~Hinton, Learning multiple layers of features from tiny
  images, Master's thesis, University of Toronto.

\bibitem{chua2009nus}
T.-S. Chua, J.~Tang, R.~Hong, H.~Li, Z.~Luo, Y.~Zheng, Nus-wide: a real-world
  web image database from national university of singapore, in: ACM
  International Conference on Image and Video Retrieval, 2009.

\bibitem{lin2014microsoft}
T.-Y. Lin, M.~Maire, S.~Belongie, J.~Hays, P.~Perona, D.~Ramanan,
  P.~Doll{\'a}r, C.~L. Zitnick, Microsoft coco: Common objects in context, in:
  European Conference on Computer Vision, 2014.

\bibitem{xiao2015learning}
T.~Xiao, T.~Xia, Y.~Yang, C.~Huang, X.~Wang, Learning from massive noisy
  labeled data for image classification, in: IEEE Conference on Computer Vision
  and Pattern Recognition, 2015.

\bibitem{huang2019coloring}
J.~Huang, M.~Patwary, G.~Diamos, Coloring big graphs with alphagozero, arXiv
  preprint:1902.10162.

\bibitem{wang2019learning}
R.~Wang, J.~Yan, X.~Yang, Learning combinatorial embedding networks for deep
  graph matching, in: IEEE International Conference on Computer Vision, 2019.

\bibitem{wang2021neural}
R.~Wang, J.~Yan, X.~Yang, Neural graph matching network: Learning lawler’s
  quadratic assignment problem with extension to hypergraph and multiple-graph
  matching, IEEE Transactions on Pattern Analysis and Machine Intelligence.

\bibitem{yan2020learning}
J.~Yan, S.~Yang, E.~R. Hancock, Learning for graph matching and related
  combinatorial optimization problems, in: International Joint Conference on
  Artificial Intelligence, 2020.

\bibitem{ma2021image}
J.~Ma, X.~Jiang, A.~Fan, J.~Jiang, J.~Yan, Image matching from handcrafted to
  deep features: A survey, International Journal of Computer Vision 129~(1)
  (2021) 23--79.

\bibitem{zaheer2017deep}
M.~Zaheer, S.~Kottur, S.~Ravanbakhsh, B.~Poczos, R.~Salakhutdinov, A.~Smola,
  Deep sets, in: Advances in Neural Information Processing Systems, 2017.

\bibitem{shen2019lorm}
Y.~Shen, Y.~Shi, J.~Zhang, K.~B. Letaief, Lorm: Learning to optimize for
  resource management in wireless networks with few training samples, IEEE
  Transactions on Wireless Communications 19~(1) (2019) 665--679.

\bibitem{jiang2017asymmetric}
Q.-Y. Jiang, W.-J. Li, Asymmetric deep supervised hashing, in: AAAI Conference
  on Artificial Intelligence, 2018.

\bibitem{liu2016towards}
H.~Liu, R.~Ji, Y.~Wu, W.~Liu, Towards optimal binary code learning via ordinal
  embedding, in: AAAI Conference on Artificial Intelligence, 2016.

\bibitem{liu2014discrete}
W.~Liu, C.~Mu, S.~Kumar, S.-F. Chang, Discrete graph hashing, in: Advances in
  Neural Information Processing Systems, 2014.

\bibitem{shen2013inductive}
F.~Shen, C.~Shen, Q.~Shi, A.~Van Den~Hengel, Z.~Tang, Inductive hashing on
  manifolds, in: IEEE Conference on Computer Vision and Pattern Recognition,
  2013.

\bibitem{tian2016global}
D.~Tian, D.~Tao, Global hashing system for fast image search, IEEE Transactions
  on Image Processing 26~(1) (2016) 79--89.

\bibitem{liu2012supervised}
W.~Liu, J.~Wang, R.~Ji, Y.-G. Jiang, S.-F. Chang, Supervised hashing with
  kernels, in: IEEE Conference on Computer Vision and Pattern Recognition,
  2012.

\bibitem{kang2016column}
W.-C. Kang, W.-J. Li, Z.-H. Zhou, Column sampling based discrete supervised
  hashing, in: AAAI Conference on Artificial Intelligence, 2016.

\bibitem{zhang2018deep}
W.~Zhang, J.~Yan, X.~Wang, H.~Zha, Deep extreme multi-label learning, in:
  International Conference on Multimedia Retrieval, 2018.

\bibitem{xu2019graph}
R.~Xu, C.~Li, J.~Yan, C.~Deng, X.~Liu, Graph convolutional network hashing for
  cross-modal retrieval, in: International Joint Conference on Artificial
  Intelligence, 2019.

\bibitem{yang2019distillhash}
E.~Yang, T.~Liu, C.~Deng, W.~Liu, D.~Tao, Distillhash: Unsupervised deep
  hashing by distilling data pairs, in: IEEE Conference on Computer Vision and
  Pattern Recognition, 2019.

\bibitem{lu2020adversarial}
J.~Lu, V.~E. Liong, Y.-P. Tan, Adversarial multi-label variational hashing,
  IEEE Transactions on Image Processing 30 (2020) 332--344.

\bibitem{sun2019supervised}
C.~Sun, X.~Song, F.~Feng, W.~X. Zhao, H.~Zhang, L.~Nie, Supervised hierarchical
  cross-modal hashing, in: International ACM SIGIR Conference on Research and
  Development in Information Retrieval, 2019.

\bibitem{li2019supervised}
C.-X. Li, T.-K. Yan, X.~Luo, L.~Nie, X.-S. Xu, Supervised robust discrete
  multimodal hashing for cross-media retrieval, IEEE Transactions on Multimedia
  21~(11) (2019) 2863--2877.

\bibitem{yan2016supervised}
T.-K. Yan, X.-S. Xu, S.~Guo, Z.~Huang, X.-L. Wang, Supervised robust discrete
  multimodal hashing for cross-media retrieval, in: ACM International on
  Conference on Information and Knowledge Management, 2016.

\bibitem{deng2019unsupervised}
C.~Deng, E.~Yang, T.~Liu, J.~Li, W.~Liu, D.~Tao, Unsupervised
  semantic-preserving adversarial hashing for image search, IEEE Transactions
  on Image Processing 28~(8) (2019) 4032--4044.

\bibitem{deng2018triplet}
C.~Deng, Z.~Chen, X.~Liu, X.~Gao, D.~Tao, Triplet-based deep hashing network
  for cross-modal retrieval, IEEE Transactions on Image Processing 27~(8)
  (2018) 3893--3903.

\bibitem{li2018deep}
N.~Li, C.~Li, C.~Deng, X.~Liu, X.~Gao, Deep joint semantic-embedding hashing,
  in: International Joint Conference on Artificial Intelligence, 2018.

\bibitem{li2020vulnerability}
C.~Li, H.~Tang, C.~Deng, L.~Zhan, W.~Liu, Vulnerability vs. reliability:
  Disentangled adversarial examples for cross-modal learning, in: ACM SIGKDD
  International Conference on Knowledge Discovery \& Data Mining, 2020.

\bibitem{li2020weakly}
Z.~Li, J.~Tang, L.~Zhang, J.~Yang, Weakly-supervised semantic guided hashing
  for social image retrieval, International Journal of Computer Vision 128.

\bibitem{jin2020deep}
L.~Jin, Z.~Li, J.~Tang, Deep semantic multimodal hashing network for scalable
  image-text and video-text retrievals, IEEE Transactions on Neural Networks
  and Learning Systems.

\bibitem{li2018deeptpami}
Z.~Li, J.~Tang, T.~Mei, Deep collaborative embedding for social image
  understanding, IEEE Transactions on Pattern Analysis and Machine Intelligence
  41~(9) (2018) 2070--2083.

\bibitem{chang2020data}
J.~Chang, Z.~Lan, C.~Cheng, Y.~Wei, Data uncertainty learning in face
  recognition, in: IEEE Conference on Computer Vision and Pattern Recognition,
  2020.

\bibitem{kendall2015bayesian}
A.~Kendall, V.~Badrinarayanan, R.~Cipolla, Bayesian segnet: Model uncertainty
  in deep convolutional encoder-decoder architectures for scene understanding,
  in: British Machine Vision Conference, 2015.

\bibitem{tang2020uncertainty}
Y.~Tang, Z.~Ni, J.~Zhou, D.~Zhang, J.~Lu, Y.~Wu, J.~Zhou, Uncertainty-aware
  score distribution learning for action quality assessment, in: IEEE
  Conference on Computer Vision and Pattern Recognition, 2020.

\bibitem{gal2016dropout}
Y.~Gal, Z.~Ghahramani, Dropout as a bayesian approximation: Representing model
  uncertainty in deep learning, in: International Conference on Machine
  Learning, 2016.

\bibitem{kendall2017uncertainties}
A.~Kendall, Y.~Gal, What uncertainties do we need in bayesian deep learning for
  computer vision?, in: Advances in Neural Information Processing Systems,
  2017.

\bibitem{wu2020unsupervised}
S.~Wu, C.~Rupprecht, A.~Vedaldi, Unsupervised learning of probably symmetric
  deformable 3d objects from images in the wild, in: IEEE Conference on
  Computer Vision and Pattern Recognition, 2020.

\bibitem{yu2019robust}
T.~Yu, D.~Li, Y.~Yang, T.~M. Hospedales, T.~Xiang, Robust person
  re-identification by modelling feature uncertainty, in: IEEE International
  Conference on Computer Vision, 2019.

\bibitem{zheng2020rectifying}
Z.~Zheng, Y.~Yang, Rectifying pseudo label learning via uncertainty estimation
  for domain adaptive semantic segmentation, International Journal of Computer
  Vision 129~(4) (2021) 1106--1120.

\bibitem{zhang2014supervised}
P.~Zhang, W.~Zhang, W.-J. Li, M.~Guo, Supervised hashing with latent factor
  models, in: International ACM SIGIR Conference on Research and Development in
  Information Retrieval, 2014.

\bibitem{lai2015simultaneous}
H.~Lai, Y.~Pan, Y.~Liu, S.~Yan, Simultaneous feature learning and hash coding
  with deep neural networks, in: IEEE Conference on Computer Vision and Pattern
  Recognition, 2015.

\bibitem{chatfield2014return}
K.~Chatfield, K.~Simonyan, A.~Vedaldi, A.~Zisserman, Return of the devil in the
  details: Delving deep into convolutional nets, in: British Machine Vision
  Conference, 2014.

\bibitem{shen2015supervised}
F.~Shen, C.~Shen, W.~Liu, H.~Tao~Shen, Supervised discrete hashing, in: IEEE
  Conference on Computer Vision and Pattern Recognition, 2015.

\bibitem{li2020general}
Q.~Li, Z.~Sun, R.~He, T.~Tan, A general framework for deep supervised discrete
  hashing, International Journal of Computer Vision 128~(8) (2020) 2204--2222.

\bibitem{zhu2016deep}
H.~Zhu, M.~Long, J.~Wang, Y.~Cao, Deep hashing network for efficient similarity
  retrieval, in: AAAI Conference on Artificial Intelligence, 2016.

\bibitem{yuan2020central}
L.~Yuan, T.~Wang, X.~Zhang, F.~E. Tay, Z.~Jie, W.~Liu, J.~Feng, Central
  similarity quantization for efficient image and video retrieval, in: IEEE
  Conference on Computer Vision and Pattern Recognition, 2020.

\end{thebibliography}

\end{document}